\documentclass[runningheads]{llncs}
\usepackage[utf8]{inputenc}
\usepackage{physics} 
\usepackage{amsmath,amsfonts,amssymb} 
\usepackage{graphicx}
\usepackage{xcolor}
\usepackage{amsfonts}
\usepackage{url}
\usepackage[export]{adjustbox}
\usepackage{subcaption}
\usepackage{graphicx}

\begin{document}
\title{What's Next? \\ Predicting Hamiltonian Dynamics from Discrete Observations of a Vector Field}

\titlerunning{Predicting Hamiltonian Dynamics in a Vector Field}

\author{Zi-Yu Khoo\inst{1} \and
Delong Zhang\inst{1} \and
Stéphane Bressan\inst{1}}
\authorrunning{Khoo et al.}

\institute{National University of Singapore. 21 Lower Kent Ridge Rd, Singapore 119077
\email{\{khoozy,zhangdel,steph\}@comp.nus.edu.sg}\\
}
\maketitle              
\begin{abstract}
We present several methods for predicting the dynamics of Hamiltonian systems from discrete observations of their vector field. Each method is either informed or uninformed of the Hamiltonian property. We empirically and comparatively evaluate the methods and observe that information that the system is Hamiltonian can be effectively informed, and that different methods strike different trade-offs between efficiency and effectiveness for different dynamical systems.
\end{abstract}

\section{Introduction}
The prediction of the dynamics of systems is a relevant and crucial task to many applications in domains ranging from the hard to social sciences. 
The dynamics of a system is captured by the vector field formed by the time derivatives of the variables describing the system's current state. The system evolves along flow lines in the state space. The flow map is the function that, given an initial state and a time interval, outputs the state of the system after the time interval.


A Hamiltonian system~\cite{Meyer1992} is a dynamical system governed by Hamilton's equations. The Hamiltonian property indicates the conservation of some quantity, typically the energy in mechanical and physical systems. 


We design, present, and evaluate physics-informed methods for the prediction of the dynamics of a system from the observation of its vector field at discrete locations of the state space. 
We want to understand and quantify the significance of informing regression and integration of a vector field with physics information. 
In the spirit of an ablation study, we compare variants of the general method and different devices, a multilayer perceptron and a Gaussian process, with and without the information that the system is Hamiltonian. In the first stage, this concerns whether the vector field is under the constraints of Hamilton's equations. In the second stage, this concerns whether the vector field is integrated with a non-symplectic or symplectic integrator. 
The empirical comparative performance evaluation is conducted with data from several physical systems of an oscillator, a pendulum, a Henon Heiles system, a Morse potential model of a dyatomic molecule, and several abstract systems with logarithmic, inverse trigonometric, exponential, radical and polynomial Hamiltonian functions.

\section{Related Work}

We are interested in the problem of learning the vector field and the flow map from samples of the vector field.
There are multiple statistical methods to learn correlated vector-valued functions~\cite{IZENMAN1975248,van1980multivariate,Wold84}. Learning vector fields using machine learning has been addressed as multiple output regression by Hastie et al.~\cite{hastie_09_statistical-learning} while state of the art works use neural networks that model ordinary differential equations~\cite{Raissi18} and partial differential equations~\cite{Ruthotto2020}. Learning vector fields using kernel methods with regularization was introduced by Micchelli et al.~\cite{Micchelli2005-hl}. Similarly, Boyle and Frean~\cite{Boyle04} introduced Gaussian processes to learn vector-valued functions. Recent works regularize the vector field~\cite{Baldassarre2010}. 

Our work is similar to that of Greydanus et al.~\cite{Greydanus_hamiltoniannn_2019}, who showed that a physics-informed neural network, similar to Bertalan et al.'s~\cite{bertalan_2019}, can faster learn the  Hamiltonian of mechanical systems and better predict their dynamics from selected samples of their vector fields than a neural network. Additionally, following Chen et al.~\cite{Chen_symplectic2020}, a symplectic integrator~\cite{Hairer2002,Vogelaere1956MethodsOI} can integrate the learned Hamiltonian vector field to predict a flow line.

\section{Methodology}
The general method for predicting the dynamics of a system from the observation of its vector field at discrete locations of its phase space comprises two successive stages: the learning or regression of the vector field from the samples, and the integration of the vector field into the flow map image of a state in the phase space for a prescribed time interval. We consider four variants of the general methods. They result from the obliviousness or awareness of information that a system is Hamiltonian during the first and second stages of the general method.

We consider two non-linear regression devices for the learning of a surrogate of the vector field, a multilayer perceptron neural network and a Gaussian process. The two devices are chosen as the main representatives of parametric and non-parametric non-linear regression statistical machine learning devices. 
They can learn a surrogate of the vector field, or 
learn a surrogate of the Hamiltonian and compute a surrogate of the vector field with automatic differentiation.

We consider the supervised learning of a surrogate $\hat{F}$ of the vector field $F$ with two physics oblivious devices: a multilayer perceptron and a Gaussian process. They regress the vector-valued function oblivious to physics information. A training data set $Z$ comprises $N$ samples, $\dv{x}{t}$ and $\dv{y}{t}$, of the vector field for $N$ states $z = (x, y)$ in the phase space of the system studied.  
When the surrogate is a multilayer perceptron, the weights minimise the mean squared error between the approximated vector field and the ground truth vector field. When the surrogate is a Gaussian process, the conditional expectation of the Gaussian process for the approximated vector field and the ground truth vector field is maximised.

We consider learning a physics-informed surrogate $\hat{H}$ of the Hamiltonian $H$ learned from the same training data set $Z$ under the constraints that the system is Hamiltonian before deriving the vector field. We adapt the method proposed by~\cite{bertalan_2019}. We  use the constraints of Equation~\ref{eqn:MLPloss1} to define the loss function of the device. $f_0$ is is an arbitrary pinning term.  $f_1$ and $f_2$ are Hamilton's equations. 
\begin{align}
    f_0 &= \left( \hat{H}(x_0,y_0) - H_0 \right)^2 \ \ &f_1 &= \left(\pdv{\hat{H}}{y} - \dv{x}{t} \right)^2& \ \ 
    f_2 &= \left(\pdv{\hat{H}}{x} + \dv{y}{t} \right)^2 
    \label{eqn:MLPloss1}
\end{align}

When the surrogate for the Hamiltonian is a multilayer perceptron neural network,
 the loss function is a linear combination of $f_0$, $f_1$ and $f_2$. 
When the surrogate is a Gaussian process, constraining its loss function leads to solving Equation~\ref{eqn:Hamiltonian_linearsystem}. In both cases, the derivative of the Hamiltonian at any new state of the phase space can be obtained from the surrogate by automatic differentiation. 
\begin{equation}
    \begin{bmatrix} \pdv{}{z_1} k(z_1,Z)^\top k(Z,Z^\prime)^{-1}\\ \vdots \\ \pdv{}{z_3} k(z_N,Z)^\top k(Z,Z^\prime)^{-1}\\  k(z_0,Z)^\top k(Z,Z^\prime)^{-1}  \end{bmatrix} [H(Z)] = \begin{bmatrix} g(Z) \\ H_0 \end{bmatrix}\label{eqn:Hamiltonian_linearsystem}
\end{equation}

The flow map for predicting the dynamics of the system is computed by integrating 
the surrogate vector fields. Ignoring that the system is Hamiltonian, one can use 
the first order explicit Euler integrator~\cite{elementaryDE_Boyce_DiPrima}. 
Knowledge of the Hamiltonian system allows use of 
the implicit symplectic Euler integrator~\cite{Hairer2002}.

\section{Performance Evaluation}
Two experiments are conducted. In the first, we empirically compare the performance of the two physics-oblivious methods learning the vector field directly and of their two physics-informed counterparts learning the Hamiltonian. We use a testing data set of $20^{2n}$ vectors at evenly spaced states in the phase space for each system. Effectiveness is measured by mean squared error between the ground truth vectors and approximated vectors, and efficiency by the time taken for early stopping of the multilayer perceptron, or the time taken to fit a Gaussian process. For the second experiment, we evaluate the prediction of the dynamics of each Hamiltonian dynamical system by computing the flow map over the interpolated vector field. The vector field are learned with physics-informed methods in the first experiment, and combined with the Euler or symplectic Euler integrator. 
We use a testing data set of $5^{2n}$ evenly spaced states in the phase space for each system. Flow lines are calculated from the differential equations of the system with a symplectic Euler integrator for 50 time steps, with step size $h=0.1$. The mean squared error between the ground truth flow line and the predicted flow line for each method, and prediction time is computed.

For both experiments, all multilayer perceptrons and physics-informed multilayer perceptrons set aside $20\%$ of the training data set for validation based early stopping.
Other settings follow Bertalan et al.~\cite{bertalan_2019}. All surrogates are trained in Google Colaboratory\footnote{Find code  and results at \url{github.com/zykhoo/predicting_hamiltonian_dynamics}.}. Training data set of size between $64$ and $1024$ are sampled uniformly at random from the vector fields of the eight example systems, respectively. The experiments are repeated for 20 unique random seeds, and the mean values are reported and compared. Results are plotted as Pareto plots.

\begin{figure}[h]
\centering
    \begin{subfigure}{0.29\textwidth}
    \includegraphics[width = \textwidth]{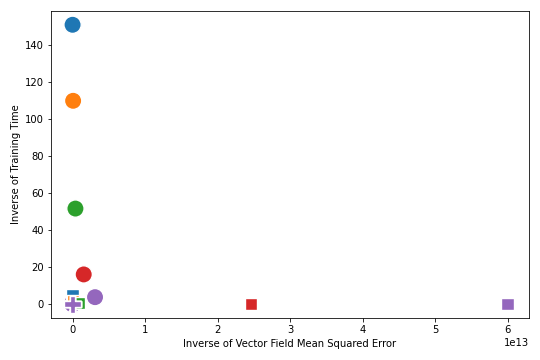}
    \caption{\tiny Simple Oscillator}
    \end{subfigure}
    \begin{subfigure}{0.29\textwidth}
    \includegraphics[width = \textwidth]{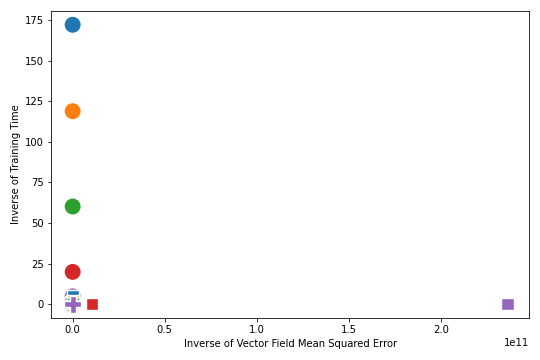}
    \caption{\tiny Nonlinear Pendulum}
    \end{subfigure}
    \begin{subfigure}{0.29\textwidth}
    \includegraphics[width = \textwidth]{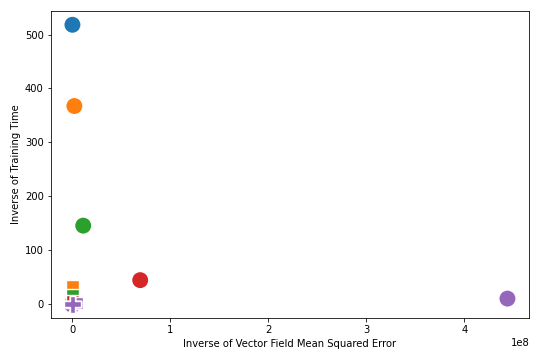}
    \caption{\tiny Henon Heiles}
    \end{subfigure}
    \begin{subfigure}{0.29\textwidth}
    \includegraphics[width = \textwidth]{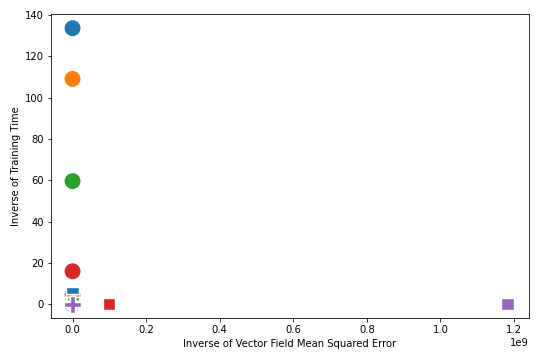}
    \caption{\tiny Morse Potential}
    \end{subfigure}
    \begin{subfigure}{0.29\textwidth}
    \includegraphics[width = \textwidth]{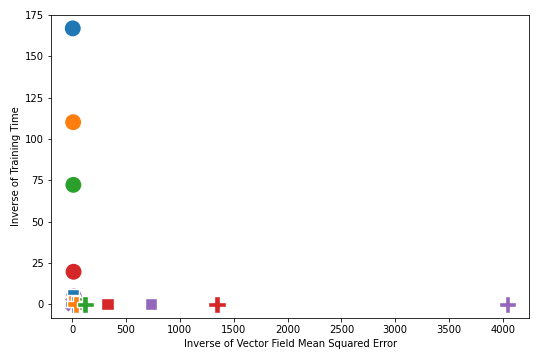}
    \caption{\tiny Logarithmic}
    \end{subfigure}
    \begin{subfigure}{0.29\textwidth}
    \includegraphics[width = \textwidth]{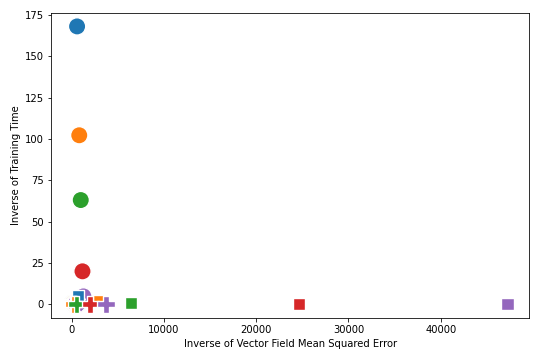}
    \caption{\tiny Inverse Cosine}
    \end{subfigure}
    \begin{subfigure}{0.29\textwidth}
    \includegraphics[width = \textwidth]{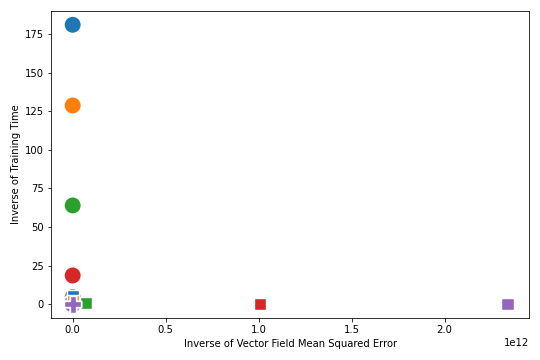}
    \caption{\tiny Exponential}
    \end{subfigure}
    \begin{subfigure}{0.29\textwidth}
    \includegraphics[width = \textwidth]{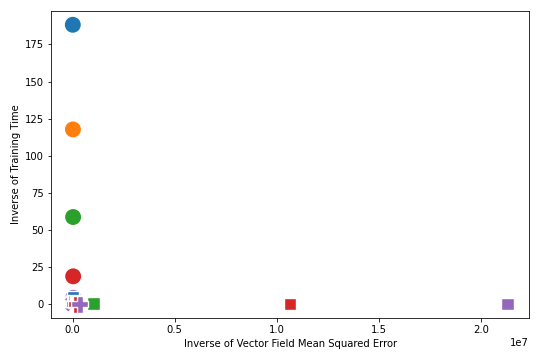}
    \caption{\tiny Square Root}
    \end{subfigure}
    \begin{subfigure}{0.29\textwidth}\centering
    \includegraphics[width =\textwidth]{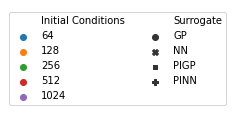}
    \caption{\tiny Legend}
    \end{subfigure}
    \caption{Learning the vector field}
    \label{fig:exp1_results}
\end{figure}

Figure~\ref{fig:exp1_results} plots the Pareto plot for the inverse of the vector field approximation mean squared error (x axis) and the inverse of surrogate training time (y axis) for the different dynamical system indicated. The colors correspond to the different training data set of varying sizes. The circle, cross, square and plus symbols represent the Gaussian process (GP), multilayer perceptron (NN), physics-informed Gaussian process (PIGP) and physics-informed multilayer perceptrons (PINN) methods respectively.  
The most efficient method is the physics-informed Gaussian process, while multilayer perceptron can better extrapolate the vector field.

\begin{figure}[h]
    \centering
    \begin{subfigure}{0.29\textwidth}
    \centering
    \includegraphics[width = \textwidth]{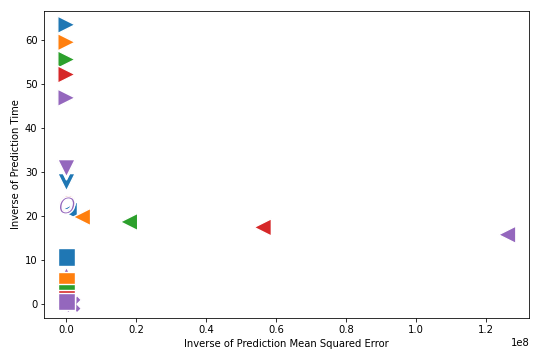}
    \caption{\tiny Simple Oscillator}
    \end{subfigure}
    \begin{subfigure}{0.29\textwidth}
    \includegraphics[width = \textwidth]{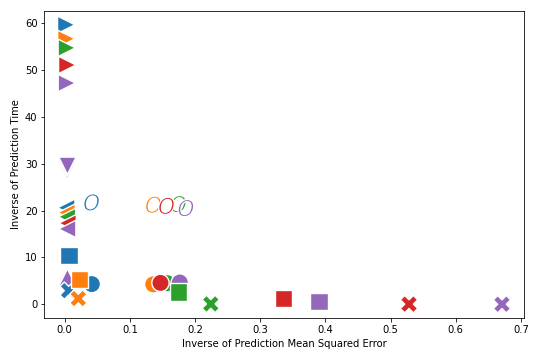}
    \caption{\tiny Nonlinear Pendulum}
    \end{subfigure}
    \begin{subfigure}{0.29\textwidth}
    \includegraphics[width = \textwidth]{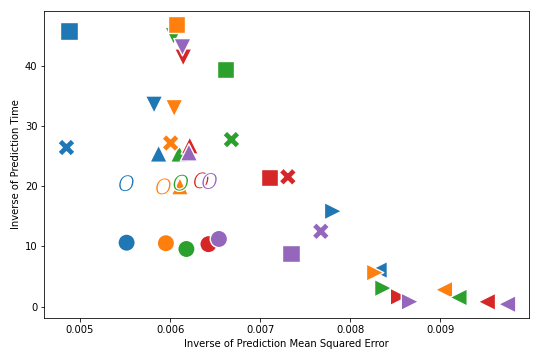}
    \caption{\tiny Henon Heiles}
    \end{subfigure}
    \begin{subfigure}{0.29\textwidth}
    \centering
    \includegraphics[width = \textwidth]{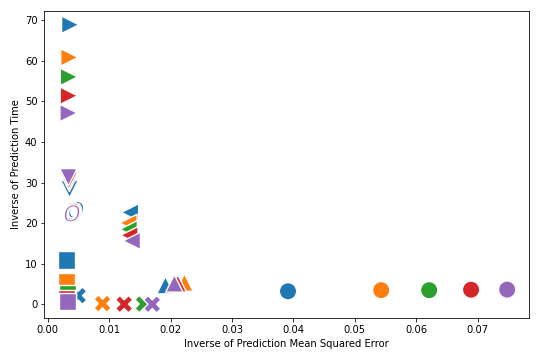}
    \caption{\tiny Morse Potential}
    \end{subfigure}
    \begin{subfigure}{0.29\textwidth}
    \includegraphics[width = \textwidth]{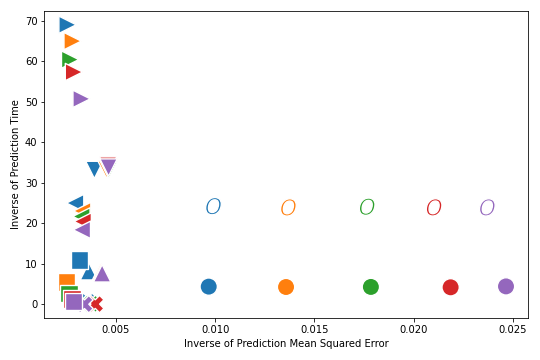}
    \caption{\tiny Logarithmic}
    \end{subfigure}
    \begin{subfigure}{0.29\textwidth}
    \includegraphics[width = \textwidth]{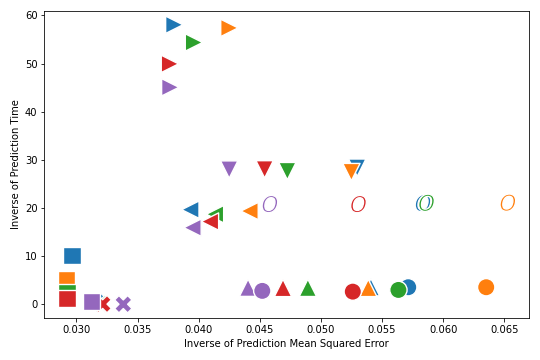}
    \caption{\tiny Inverse Cosine}
    \end{subfigure}
    \begin{subfigure}{0.29\textwidth}
    \centering
    \includegraphics[width = \textwidth]{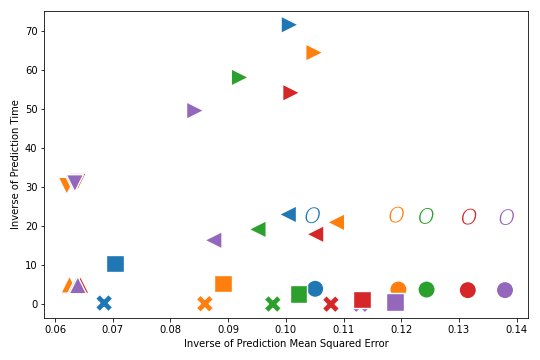}
    \caption{\tiny Exponential}
    \end{subfigure}
    \begin{subfigure}{0.29\textwidth}
    \includegraphics[width = \textwidth]{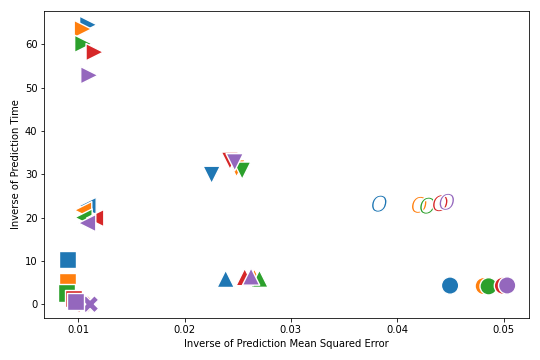}
    \caption{\tiny Square Root}
    \end{subfigure}
    \begin{subfigure}{0.29\textwidth} \centering
    \includegraphics[width = 0.7\textwidth]{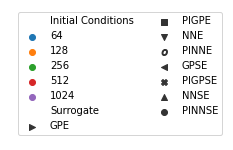}
    \caption{\tiny Legend}
    \end{subfigure}
    \caption{Predicting the flow map}
    \label{fig:exp2_results}
\end{figure}

Figure~\ref{fig:exp2_results} compares inverse prediction error (x axis) and inverse prediction time (y axis). The colors correspond to training data set of varying sizes. The right-pointing triangle, square, down-pointing triangle and empty circle symbols represent the Gaussian process (GPE), physics-informed Gaussian process (PIGPE), multilayer perceptron (NNE), and physics-informed multilayer perceptron (PINNE), all with the Euler integrator. The left-pointing triangle, cross, up-pointing triangle and filled circle symbols represent the same surrogates with symplectic Euler integrator. They learn and integrate the vector field.
The most efficient method for the prediction of the dynamics is the physics-informed multilayer perceptron with symplectic integrator. This demonstrates the advantage of informing the methods of the Hamiltonian nature of the dynamical systems 

\section{Conclusion}
We design, present, and evaluate physics-informed methods for the prediction of the dynamics of a system from the observation of its vector field at discrete locations of the state space. We show that information that the system is Hamiltonian can be effectively informed in both the regression and integration of the vector field. 
The methods strike trade-offs between efficiency and effectiveness. 


\section*{Acknowledgement} This research is partially supported by the Agency of Science, Technology and Research (A*STAR), by the National Research Foundation, Prime Minister’s Office, Singapore, under its Campus for Research Excellence and Technological Enterprise (CREATE) programme as part of the programme DesCartes, and by the Ministry of Education, Singapore, under its Academic Research Fund Tier 2 grant call (Award ref:  MOE-T2EP50120-0019). Any opinions, findings and conclusions or recommendations expressed in this material are those of the authors and do not reflect the views of the National Research Foundation or of the Ministry of Education, Singapore.

\bibliographystyle{splncs04}
\bibliography{cite}

\end{document}